\pdfoutput=1
\documentclass[11pt]{article}

\usepackage[final]{acl}

\usepackage{times}
\usepackage{latexsym}

\usepackage[T1]{fontenc}
\usepackage[utf8]{inputenc}

\usepackage{microtype}

\usepackage{inconsolata}

\usepackage{graphicx}
\usepackage{makecell}

\title{Multi-document Summarization through Multi-document Event Relation Graph Reasoning in LLMs: a case study in Framing Bias Mitigation}

\author{Yuanyuan Lei and Ruihong Huang\\
        Department of Computer Science and Engineering\\
        Texas A\&M University, College Station, TX\\
        \texttt{\{yuanyuan, huangrh\}@tamu.edu}}

\begin{document}
\maketitle
\begin{abstract}

Media outlets are becoming more partisan and polarized nowadays. Most previous work focused on detecting media bias. In this paper, we aim to mitigate media bias by generating a neutralized summary given multiple articles presenting different ideological views. Motivated by the critical role of events and event relations in media bias detection, we propose to increase awareness of bias in LLMs via multi-document events reasoning and use a multi-document event relation graph to guide the summarization process. This graph contains rich event information useful to reveal bias: four common types of in-doc event relations to reflect content framing bias, cross-doc event coreference relation to reveal content selection bias, and event-level moral opinions to highlight opinionated framing bias. We further develop two strategies to incorporate the multi-document event relation graph for neutralized summarization. Firstly, we convert a graph into natural language descriptions and feed the textualized graph into LLMs as a part of a hard text prompt. Secondly, we encode the graph with graph attention network and insert the graph embedding into LLMs as a soft prompt. Both automatic evaluation and human evaluation confirm that our approach effectively mitigates both lexical and informational media bias, and meanwhile improves content preservation\footnote{The code and data link is: \url{https://github.com/yuanyuanlei-nlp/multi_doc_summarization_acl_2025}}.

\end{abstract}

\section{Introduction}

\begin{figure*}[t]
  \centering
  \includegraphics[width = 6.3in]{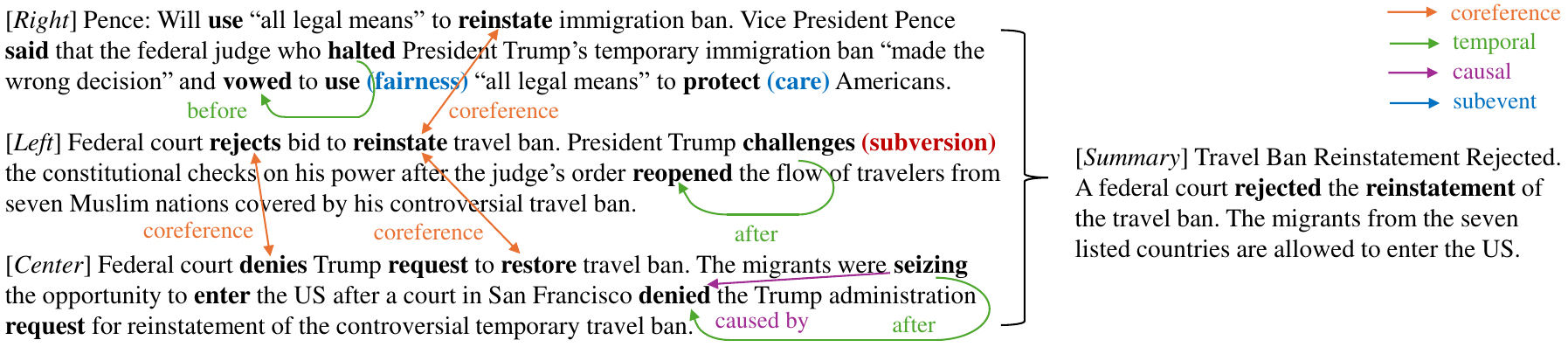}
  \caption{An example of multi-document event relation graph based on a triplet of articles. The multi-document event relation graph consists of events as nodes (bold words), moral opinions as event node attribute (colored words in parentheses), and within-doc and cross-doc event relations (colored edges between events).}
  \label{introduction_example}
\end{figure*}

Media bias refers to the practice of presenting biased or partial information in news articles to promote an ideological leaning and sway readers' political opinions \cite{gentzkow2006media, groeling2013media, morstatter2018identifying}. News media plays a crucial role not only in supplying information, but also in selecting and organizing information to shape public opinions \cite{baron2005competing, asp2007fairness, hillebrand2019role, lei-huang-2022-shot, lei-etal-2024-polarity}. With the media outlets become more partisan and polarized, the journalists usually embed their ideological bias into news articles through content framing \cite{tankard2001empirical, prior2013media, d2017framing}. The prevalence of media bias has harmful effects on both individuals and society, such as misleading audiences, intensifying societal polarization, and undermining democratic values \cite{kuypers2002press, druckman2005impact, 10008998, entman2007framing, emami2020impact}.

%While media framing bias presents a significant issue, most previous research focused on detecting framing bias, and few efforts were made to mitigate framing bias \cite{fan-etal-2019-plain, baly2020we, naredla2022detection, liu2022quantifying}. In this paper, we aim to mitigate framing bias by generating a neutralized summary from multiple articles with diverse ideology. Specifically, given a cluster of news articles that frame the same story from liberal or conservative viewpoints, we aim to provide the readers with an impartial and unbiased summary, with media framing bias mitigated \cite{lee-etal-2022-neus}. This task of neutralized summarization filters out ideological bias in content framing, offers a comprehensive view of news reporting, and enables objective information access.

While media bias presents a significant issue, most previous research focused on detecting media bias and few efforts were made to mitigate media bias \cite{fan-etal-2019-plain, baly2020we, naredla2022detection, liu2022quantifying}. Recently, neutralized summarization \cite{lee-etal-2022-neus} was proposed to mitigate media bias by generating a neutralized summary given multiple articles that frame the same story from liberal or conservative viewpoints. This task holds great promise in offering a comprehensive view of news reporting and enabling unbiased information access. However, the approaches for neutralized summarization remain rudimentary and mainly rely on basic text-to-text generation that may suffer from limited content analysis or lack of awareness of bias \cite{lee-etal-2022-neus, bang-etal-2023-mitigating}.

To mitigate media bias, we argue that it is necessary to incorporate bias indicators to inform the neutralized summarization process. Motivated by the critical roles of events and event relations in detecting media bias \cite{zhang-etal-2021-salience, liu-etal-2023-things, zhang2024moka, lei-huang-2024-sentence}, we propose to inform LLMs of bias distribution through multi-document events reasoning.

In particular, we propose to build a multi-document event relation graph that takes events as nodes and captures in-doc and cross-doc event relations. This graph contains various bias relevant information: (1) four types of in-doc event relations (temporal, causal, subevent, and coreference) illustrate the diverse narrative logic connecting events within an article, thereby reflecting content framing bias (2) cross-doc event coreference relations distinguish between events commonly reported across multiple articles and events selectively reported by a particular article, thus revealing content selection bias (3) we also add event-level moral opinions as a feature for event nodes to highlight %opinionated framing bias 
morally opinionated events 
and indicate moral framing differences across articles. The designed multi-document event relation graph is expected to increase awareness of bias in LLMs and serve as an useful guidance for neutralized summarization.

Take the example in Figure \ref{introduction_example} as an illustration, where the three articles report essentially the same news of {\it Travel Ban Reinstatement}, but from perspectives of different ideological leanings. The bold words are event words and the edges are colored to represent different event relation types. We can see that in this multi-document event relation graph, the cross-doc event coreference relations highlight common events shared across the articles, and the main {\it Reinstatement} event appeared in all the articles. Then, among the remaining events unique to each article, the right leaning article chooses to frame the main event as {\it use "all legal means"} and describe the purpose as {\it protect Americans}, while the left leaning article frames the main event as {\it challenges the constitutional checks}. Further, moral sentiment analysis shows that {\it use "all legal means"} and {\it protect Americans} have positive moral sentiments of fairness and care respectively, while {\it challenges the constitutional checks} has a negative moral sentiment of subversion.

We further propose a framework to integrate the multi-document event relation graph into LLMs for neutralized summarization, which consists of two key components. The first is graph textualization, where we convert the multi-doc event relation graph into natural language descriptions, and feed the textualized graph as a hard prompt into LLMs. The second is graph prompt tuning, where we encode the multi-doc event relation graph with graph attention network, and insert the graph embedding as a soft prompt into LLMs for tuning. The incorporation of hard and soft prompt are complementary: the hard prompt informs the model of graph structure by augmenting the instruction, while the soft prompt enables direct tuning on graph embedding. Both automatic evaluation and human evaluation confirm the effectiveness of our approach based on multi-doc event relation graph, which notably mitigates both lexical and informational media bias in summaries and meanwhile improves the preservation of content semantics. Our main contributions are summarized as follows:

\begin{itemize}
    %\item We demonstrate the pivotal role of events and event relations in mitigating framing bias.
    %\item We initially propose a multi-document event relation graph to guide neutral summarization.
    %\item We design a new framework to integrate the graph into LLMs, reducing framing bias while preserving content semantics on NeuS dataset.
    \item We propose to incorporate bias relevant information for media bias mitigation through multi-document events reasoning.
    \item We introduce a multi-document event relation graph to guide neutralized summarization.
    \item We design a new framework to integrate the graph into LLMs, reducing media bias while also improving content preservation.
\end{itemize}

\section{Related Work}

\noindent\textbf{Multi-document Summarization} aims to generate a concise and informative summary from a collection of documents \cite{lebanoff-etal-2018-adapting}. In recent years, researchers applied deep neural networks and large pre-trained language models for multi-document summarization \cite{mao-etal-2020-multi, pasunuru-etal-2021-efficiently, shen-etal-2023-hierarchical}. Additionally, researchers explored several subtopics in this field, such as topic-guided, agreement-oriented, or entity-aware summarization \cite{cui-hu-2021-topic-guided, pang-etal-2021-agreesum, zhou-etal-2021-entity}. Differently, our goal is to generate a neutralized and unbiased summary from multiple articles with varying ideology, thereby mitigating framing bias.

\noindent\textbf{Event Graph} was introduced by \citet{li-etal-2020-connecting, jin-etal-2022-event}, which includes entity-entity links and event-entity links via event argument roles, yet lacks event-event relations. This graph is employed in several down-stream tasks, including story generation \cite{chen-etal-2021-graphplan}, misinformation detection \cite{wu-etal-2022-cross}, and sentence fusion \cite{yuan-etal-2021-event}. In contrast, our graph focuses on events and establishes interrelations between events through four types of event-event relations.

\noindent\textbf{Event Relations} were studied for decades. There are four common relations between events: coreference, temporal, causal, and subevent relations \cite{caselli-vossen-2017-event, zeng-etal-2020-event, tan-etal-2021-extracting, man-etal-2022-event, lai-etal-2022-multilingual-subevent, wang-etal-2022-maven, lei-huang-2023-identifying}. While each type of relations was previously studied in isolation, we aim to develop a model that unifies all four relations for comprehensive content analysis. Instead of analyzing event relations in single article, we propose to construct a multi-document event relation graph to capture narrative structures across various articles.

\noindent\textbf{Media Bias Detection} attracted research interests for years \cite{lichter2017theories}. Early work detect media bias at source level, by assuming all the articles within one media source share the same ideology \cite{budak2016fair, baly2018predicting}. Subsequent research shifted towards detecting media bias at article level, by classifying the ideology leaning of each article \cite{sapiro2019examining, baly2020we, chen2020detecting, liu-etal-2022-politics}. More recently, there has been an interest in detecting media bias at more granular levels, such as sentence level or token level \cite{da-san-martino-etal-2019-fine, van-den-berg-markert-2020-context, spinde-etal-2021-neural-media, vargas-etal-2023-predicting, lei-etal-2022-sentence, lei-huang-2023-discourse, lei-huang-2024-boosting}. Different from most previous work that develop approaches for media bias detection, this paper aims for media bias mitigation.

\noindent\textbf{Media Bias Mitigation} has a relatively short research history. The first work for mitigating media framing bias was introduced by \citet{lee-etal-2022-neus}, where they introduced the NeuS dataset for neutralized summarization. They also designed a method to generate summary in a hierarchical order from title to article \cite{lee-etal-2022-neus}. Different from previous work based on text-to-text model, our work firstly incorporates bias information into this task. We propose a multi-document event relation graph approach to inform LLMs of bias distribution and guide LLMs in mitigating framing bias.

\section{Multi-document Event Relation Graph}

Given a cluster of news articles, we propose to create a multi-document event relation graph for content analysis. Overall, the graph comprises events as nodes, moral opinions as event node attributes, four common types of single-doc event relations, as well as cross-doc event coreference relation to connect articles together.

\subsection{Event and Moral Attributes}

An \textit{event} refers to an occurrence or action reported in news articles, and is the basic element in story telling \cite{ogorman-etal-2016-richer}. In news media, the authors often convey their political stance through moral judgment towards events, evaluating whether the events align with social moral rules. The sociologists have developed Moral Foundation Theory to categorize social moral rules into five dimensions, each associated with a positive and negative judgment: \textit{Care / Harm}, \textit{Fairness / Cheating}, \textit{Loyalty / Betrayal}, \textit{Authority / Subversion}, and \textit{Purity / Degradation} \cite{graham2009liberals}.

The first step of graph construction is extracting events from each article. An event identification model is trained on the MAVEN dataset which annotated event mentions for general-domain documents \cite{wang-etal-2020-maven} (details in Appendix \ref{eval_graph}). Given a candidate article consisting of $N$ words, we infer the trained event identifier to predict the probability of each word triggering an event:
\begin{equation}
    P_{i}^{event} = (p_i^{event}, p_i^{non-event})
\end{equation}
Subsequently, we extract the moral opinion towards each event as node attribute. A moral classifier is trained based on the EMONA dataset which annotated event-level moral opinions in news articles \cite{lei2024emona}. For all the extracted events, we use the moral classifier to predict their moral judgments into ten moral values or non-moral class:
\begin{equation}
    P_i^{moral} = (p_i^{care}, p_i^{harm}, ..., p_i^{non-moral})
\end{equation}

\begin{figure*}[t]
  \centering
  \includegraphics[width = 6.3in]{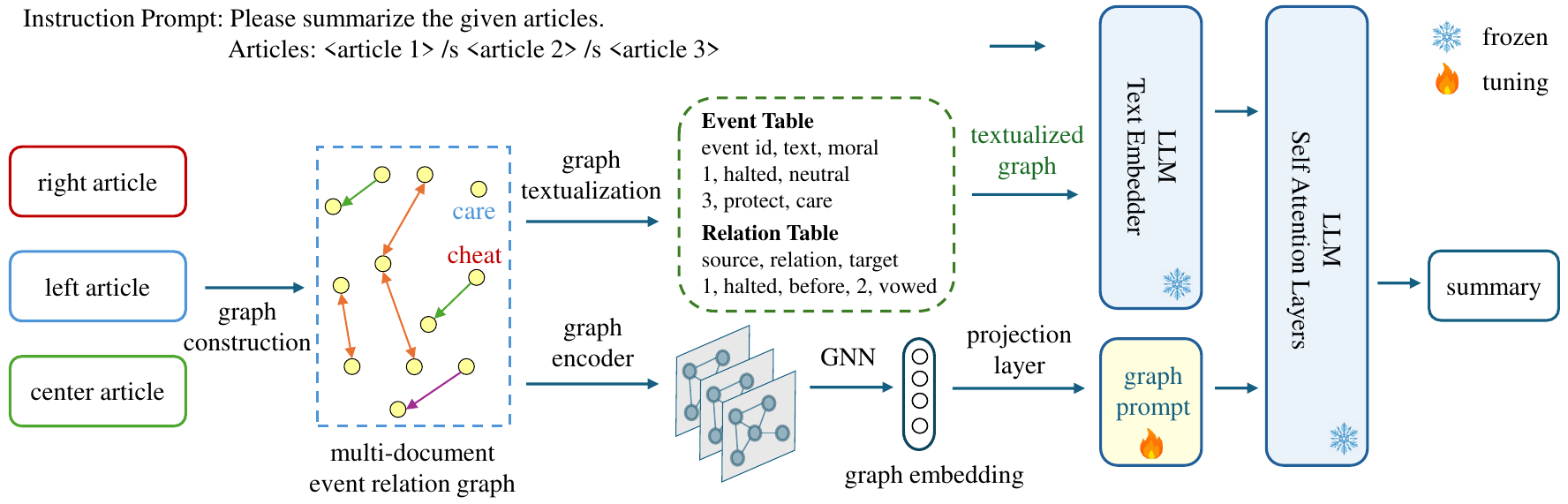}
  \caption{An illustration of neutralized summarization guided by multi-document event relation graph.}
  \label{methodology_figure}
\end{figure*}

\subsection{Event-Event Relations}

There are four common event relations. Coreference relation informs us whether the two events designate the same occurrence or not. Temporal relation represents the chronological orders between events, such as \textit{before}, \textit{after}, and \textit{overlap}. Causal relation shows the causality or precondition relation between events, and is categorized into \textit{causes} and \textit{caused by}. Subevent relation recognizes containment or subordination relation between events, including \textit{contains} and \textit{contained by} classes.

The next step is connecting the events with four event relations for each article. The four event relation extractors are trained on the general-domain MAVEN-ERE dataset \cite{wang-etal-2022-maven}. Since the four event relations interact with each other to form a cohesive narrative structure, we adopt the joint learning framework to train these relations collaboratively \cite{wang-etal-2022-maven}. During the inference process, we establish all possible event pairs based on the extracted events. For each event pair $(event_i, event_j)$, we employ the trained relations extractors to predict the probabilities for the four relations. The final label for each relation is derived by applying the $argmax$ function on these predicted probabilities:
\begin{equation}
    P_{i, j}^{corefer} = (p_{i, j}^{corefer}, p_{i, j}^{non-corefer})
\end{equation}
\begin{equation}
    P_{i, j}^{tem} = (p_{i, j}^{before}, p_{i, j}^{after}, p_{i, j}^{overlap}, p_{i, j}^{non-tem})
\end{equation}
\begin{equation}
    P_{i, j}^{causal} = (p_{i, j}^{causes}, p_{i, j}^{caused-by}, p_{i, j}^{non-causal})
\end{equation}
\begin{equation}
    P_{i, j}^{sub} = (p_{i, j}^{contains}, p_{i, j}^{contained-by}, p_{i, j}^{non-sub})
\end{equation}

\subsection{Cross-doc Event Coreference}

The final step is connecting the events across different documents through event coreference relation. We employ a cross-document event coreference resolution system \cite{lai-etal-2021-context, wen-etal-2021-resin} to identify clusters of events from multiple documents. The cross-document event coreference relation is used to connect multiple single-document event relation graphs together, facilitating cross-document content analysis and narrative comparison.

\section{Neutralized Summarization}

The multi-document event relation graph is then incorporated into LLMs for neutralized summarization through two key components (Figure \ref{methodology_figure}). The first is graph textualization, where we convert the graph into natural language descriptions, and feed the textualized graph into LLMs as a hard prompt. The second is graph prompt tuning, where we encode the graph with graph neural network, and insert the graph embedding into LLMs as a soft prompt. The hard and soft prompts complement each other: the hard prompt augments the instruction with graph structure, while the soft prompt enables direct tuning on graph embeddings.

\subsection{Graph Textualization}

The graph textualization process is designed to transform the graph structure into a natural language format, making it readable by LLMs. This involves creating an event table $T_{event}$ to describe the events information, including event id, event text, and event-level moral judgment. Additionally, a relation table $T_{relation}$ is developed to describe the relations information between events, which includes columns for source event, relation, and target event. The two tables $T_{event}$ and $T_{relation}$ convert the graph structure into textual descriptions, resulting in a textualized graph. This textualized graph is then fed into LLMs as a hard prompt:
\begin{equation}
    h_t = TextEmbedder(T_{event}; T_{relation})
\end{equation}
where $TextEmbedder$ is the text embedding layer of a pre-trained and frozen LLM.

\subsection{Graph Prompt Tuning}

The graph prompt tuning process is designed to create a graph embedding and project it as a soft prompt into LLMs for further tuning. This involves a graph propagation process to update events embeddings with their neighbor events embeddings through interconnected relations, and produce a final graph embedding that represents the entire graph. In addition, a projection layer is crafted to transform the graph embedding into the same representation space of LLMs.

During the graph propagation process, we encode the article with Longformer \cite{beltagy2020longformer}, and use the corresponding word embeddings to initialize event node embeddings $e_i$. Then, we update event embeddings with their moral values:
\begin{equation}
    e_i = W^m (e_i \oplus m_i) + b^m
\end{equation}
where $m_i$ is the moral label embedding of the event $e_i$, $\oplus$ denotes feature concatenation, $W^m, b^m$ are trainable parameters of a transformation layer.

Afterwards, we develop a relation-aware graph attention network to update event embeddings with neighbor events embeddings through their linked relations. Given a pair of events $(e_i, e_j)$, their relation $r_{ij}$ is initialized as the embedding of the corresponding relation word. At the $l$-th layer, the input for $i$-th event node are output features produced by the previous layer denoted as $e_i^{(l-1)}$. The relation embedding $r_{ij}$ is updated as:
\begin{equation}
    r_{ij} = W^r (e_i^{(l-1)} \oplus r_{ij} \oplus e_j^{(l-1)})
\end{equation}
where $W^r$ are trainable matrix. Then the attention weights across neighbor events are computed as:
\begin{equation}
    \alpha_{ij} = softmax_j \Bigl((W^Q e_i^{(l-1)})(W^K r_{ij})^T\Bigr)
\end{equation}
where $W^Q$, $W^K$ are trainable parameters. The output feature for $e_i$ regarding the relation $r$ is :
\begin{equation}
    e_{i,r}^{(l)} = \sum_{j\in \mathcal{N}_{i,r}} \alpha_{ij} W^V r_{ij}
\end{equation}
where $\mathcal{N}_{i,r}$ denotes the neighbor event nodes that connect with event $e_i$ via the relation $r$, and $r \in R = $ \{\textit{coreference}, \textit{before}, \textit{after}, \textit{overlap}, \textit{causes}, \textit{caused by}, \textit{contains}, \textit{contained by}\}. After collecting $e_{i,r}^{(l)}$ for all relation types $R$, the final output feature for event $e_i$ at $l$-th layer is aggregated as:
\begin{equation}
    e_i^{(l)} = \sum_{r \in R} e_{i,r}^{(l)} / |R|
\end{equation}

Subsequently, we derive the graph embedding by introducing an additional graph node and linking it to the event nodes. We employ the standard graph attention network to aggregate event embeddings into the graph embedding $h_g^{l}$ at $l$-th layer:
\begin{equation}
    \alpha_i = softmax_i \Bigl(W h_g^{(l-1)} \oplus W e_i^{(l-1)}\Bigr)
\end{equation}
\begin{equation}
    h_g^{(l)} = \sum_i \alpha_i W e_i^{(l-1)}
\end{equation}
The graph embedding from the last layer is designated as the final graph embedding $h_g$.

Furthermore, a projection layer is designed to transform the graph embedding into the same representation space of the LLM:
\begin{equation}
    \hat{h}_g = W_2 \Bigl(W_1 h_g + b_1\Bigr) + b_2
\end{equation}
where $W_1$, $W_2$, $b_1$, $b_2$ are the parameters of the projection layer, and $\hat{h}_g$ is the resulting graph prompt.

During the summarization procedure, both the graph prompt $\hat{h}_g$ and textualized graph $h_t$ are fed into the self attention layers of a pre-trained and frozen LLM. The graph prompt $\hat{h}_g$ receives gradients and enables back propagation.

\begin{table*}[t]
    \centering
    \scalebox{0.74}{\begin{tabular}{|l||ccccc||cccc|}
        \hline
        & \multicolumn{5}{c||}{Content Evaluation} & \multicolumn{4}{c|}{Bias Evaluation} \\
        & Rouge-1 & Rouge-2 & Rouge-L & Rouge-Lsum & BLEU-2 & polarization & p-arousal & n-arousal & sum-arousal \\
        \hline
        Baselines & & & & & & & & & \\
        LexRank & 42.24 & 18.16 & 26.61 & 36.87 & 17.68 & 63.73 & 2.42 & 1.53 & 3.95 \\
        BART-CNN & 38.22 & 15.73 & 25.52 & 34.25 & 15.26 & 76.67 & 2.02 & 1.14 & 3.16 \\
        BART-Multi & 39.60 & 16.26 & 24.57 & 35.05 & 17.60 & 54.47 & 3.46 & 1.64 & 5.10 \\
        Pegasus-CNN & 38.17 & 15.61 & 25.18 & 31.17 & 13.69 & 74.45 & 1.98 & 1.15 & 3.12 \\
        Pegasus-Multi & 35.33 & 13.27 & 21.25 & 31.14 & 14.79 & 59.15 & 4.83 & 2.27 & 7.10 \\
        GPT-3.5 & 42.01 & 16.25 & 26.13 & 37.27 & 18.77 & 77.25 & 3.40 & 2.13 & 5.52 \\
        GPT-3.5 + one-shot & 41.95 & 16.77 & 28.13 & 37.39 & 18.33 & 44.29 & 2.52 & 1.57 & 4.08 \\
        GPT-3.5 + graph & 43.20 & 18.06 & 30.56 & 38.08 & 18.99 & 32.37 & 2.03 & 1.59 & 3.62 \\
        GPT-4 & 42.36 & 16.49 & 26.30 & 37.31 & 19.04 & 75.86 & 3.37 & 1.97 & 5.34 \\
        GPT-4 + graph & 42.61 & 18.67 & 30.82 & 38.18 & 19.09 & 31.77 & 2.11 & 1.49 & 3.60 \\
        NeuS & 39.09 & 18.93 & 29.74 & 35.35 & 16.21 & 38.51 & 1.69 & 0.83 & 2.53 \\
        \citet{bang-etal-2023-mitigating} & - & - & - & - & - & - & 1.57 & 0.91 & 2.48 \\
        \hline
        LED & 40.30 & 18.63 & 30.24 & 36.26 & 17.30 & 31.97 & 1.59 & 0.86 & 2.45 \\
        + textual graph & 41.84 & 19.46 & 31.18 & 37.30 & 18.23 & 29.77 & 1.29 & 0.83 & 2.12 \\
        + graph prompt & 42.06 & 19.96 & 31.74 & 37.56 & 18.23 & 29.84 & 1.33 & 0.84 & 2.17 \\
        + both (full model) & 42.96 & 20.66 & 32.74 & 38.56 & 19.09 & 28.14 & \textbf{1.26} & \textbf{0.71} & \textbf{1.97} \\
        \hline
        Llama-2 & 42.26 & 19.25 & 30.88 & 37.75 & 19.15 & 30.30 & 1.80 & 1.02 & 2.81 \\
        + textual graph & 43.98 & 20.52 & 32.57 & 39.30 & 20.18 & 28.22 & 1.59 & 0.94 & 2.53 \\
        + graph prompt & 44.44 & 21.01 & 32.95 & 39.97 & 20.42 & 28.01 & 1.50 & 1.01 & 2.50 \\
        + both (full model) & \textbf{45.14} & \textbf{22.30} & \textbf{34.02} & \textbf{40.74} & \textbf{21.89} & \textbf{27.89} & 1.55 & 0.90 & 2.46 \\
        \hline
    \end{tabular}}
    \caption{Automatic Evaluation results of neutralized summarization on NeuS dataset. 
   We calculate the Rouge and BLEU scores to evaluate content preservation; and we calculate the polarization score and the arousal scores to evaluate content-level informational bias and lexical-level bias respectively. 
   The summarizer with better performance should attain higher Rouge and BLEU scores, but lower polarization score and arousal scores.}
    \label{automatic_evaluation}
\end{table*}

\section{Experiments}

\subsection{Datasets}

The task of neutralized summarization has a relatively short research history, and NeuS \cite{lee-etal-2022-neus} is the only available dataset up till now.

\noindent\textbf{NeuS} \cite{lee-etal-2022-neus} collects US political news articles from AllSides website. The articles that discuss the same event and present different ideological views are grouped together as a cluster. Each cluster contains three articles, and each article comes from liberal, center, conservative media sources respectively. The dataset also provides an expert written summary for each cluster of articles. We follow the dataset splitting released by \citet{lee-etal-2022-neus}, which results in 2452 / 307 / 307 news clusters allocated to the train, valid, test sets.

\subsection{Experimental Settings}

To validate our approach, we use two types of language models as the foundation model for summarization in the experiments: a decoder-only model and a encoder-decoder model. For the decoder-only model, we choose the open-source large language model LLama-2 and use the version of llama-2-7b-chat-hf \cite{touvron2023llama}. For the encoder-decoder model, considering the input text is typically long, we choose the longformer encoder-decoder (LED) model and use the version of led-large-16384 \cite{beltagy2020longformer}.

The models take the instruction prompt and a cluster of three articles as input, and generates a summary as output. The instruction prompt provided to the Llama-2 and LED models is detailed in Appendix \ref{method_prompt}. The maximum input length is set as 2048, maximum output length is 512, number of epochs is 5, gradient accumulation step is 16, weight decay is 1e-2, learning rate for Llama-2 is 1e-4 and learning rate for LED is 1e-5. The Llama-2 model is trained with LoRA \cite{hu2021lora}, with the rank 8, alpha 16 and dropout 0.05.

\begin{table*}[t]
    \centering
    \scalebox{0.8}{\begin{tabular}{|l||cc||cc||cc|}
        \hline
        & \multicolumn{2}{c||}{Bias Evaluation} & \multicolumn{2}{c||}{Content Evaluation} & \multicolumn{2}{c|}{Language Evaluation} \\
        & Lexical Bias & Informational Bias & Non-Hallucination & Recovery & Fluency & Coherency \\
        \hline
        NeuS & 85.90 & 88.46 & 71.79 & 68.42 & 61.54 & 76.92 \\
        GPT-4 & 80.77 & 79.49 & \textbf{89.74} & \textbf{97.43} & 97.43 & \textbf{97.43} \\
        LED & 81.58 & 78.20 & 73.68 & 73.68 & 84.21 & 89.74 \\
        LED + graph & \textbf{92.10} & 85.90 & 85.00 & 78.95 & 89.47 & 92.31 \\
        Llama-2 & 83.33 & 84.61 & 68.42 & 74.36 & \textbf{100.00} & 94.87 \\
        Llama-2 + graph & 91.02 & \textbf{89.74} & 84.21 & 87.18 & \textbf{100.00} & \textbf{97.43} \\
        \hline
    \end{tabular}}
    \caption{Human Evaluation results of neutralized summarization on NeuS dataset. The bias evaluation includes both lexical bias and informational bias. The higher scores for the six metrics represent better performance. The row "+ graph" means incorporating the multi-doc event relation graph.}
    \label{human_evaluation}
\end{table*}

\begin{table*}[t]
    \centering
    \scalebox{0.72}{\begin{tabular}{|l||ccccc||cccc|}
        \hline
        & \multicolumn{5}{c||}{Content Evaluation} & \multicolumn{4}{c|}{Bias Evaluation} \\
        & Rouge-1 & Rouge-2 & Rouge-L & Rouge-Lsum & BLEU-2 & polarization & p-arousal & n-arousal & sum-arousal \\
        \hline
        Llama-2 & 42.26 & 19.25 & 30.88 & 37.75 & 19.15 & 30.30 & 1.80 & 1.02 & 2.81 \\
        + event moral & 43.82 & 20.65 & 32.48 & 39.42 & 20.11 & 29.05 & 1.58 & 0.93 & 2.51 \\
        + in-doc relations & 44.74 & 21.31 & 33.11 & 40.22 & 21.01 & 28.57 & 1.68 & 1.00 & 2.68 \\
        + cross-doc coreference & 44.53 & 20.78 & 32.80 & 39.53 & 20.72 & 28.16 & 1.63 & 0.97 & 2.60 \\
        + all (full model) & \textbf{45.14} & \textbf{22.30} & \textbf{34.02} & \textbf{40.74} & \textbf{21.89} & \textbf{27.89} & \textbf{1.55} & \textbf{0.90} & \textbf{2.46} \\
        \hline
    \end{tabular}}
    \caption{The ablation study of different components in the multi-document event relation graph. The summarizer with better performance should attain higher Rouge and BLEU scores, lower polarization score and arousal scores.}
    \label{ablation_study}
\end{table*}

\subsection{Baselines}

We implemented baselines that are mentioned in \citet{lee-etal-2022-neus} for comparison. Furthermore, we also establish several GPT-based baselines.

\vspace{2pt}

\noindent\textbf{LexRank} \cite{Erkan_2004} is an unsupervised model that selects sentences based on graph centrality and generates extractive summaries.

\vspace{2pt}

\noindent\textbf{BART-CNN} \cite{lewis-etal-2020-bart} is a news summarization model that fine tunes BART-large on the CNN Daily Mail dataset \cite{nallapati-etal-2016-abstractive}.

\vspace{2pt}

\noindent\textbf{BART-Multi} \cite{lewis-etal-2020-bart} is a multi-news summarization model that fine tunes BART-large on the Multi-News dataset \cite{fabbri-etal-2019-multi}.

\vspace{2pt}

\noindent\textbf{Pegasus-CNN} \cite{zhang2020pegasus} is a news summarization model that fine tunes Pegasus-large on the CNN Daily Mail dataset \cite{nallapati-etal-2016-abstractive}.

\vspace{2pt}

\noindent\textbf{Pegasus-Multi} \cite{zhang2020pegasus} is a multi-news summarization model that fine tunes Pegasus-large on the Multi-News dataset \cite{fabbri-etal-2019-multi}.

\vspace{2pt}

\noindent\textbf{NeuS} \cite{lee-etal-2022-neus} develops an abstractive summarization method that learns to generate summary in a hierarchical order from title to article.

\vspace{2pt}

\noindent\citet{bang-etal-2023-mitigating} designs a polarity minimization loss function to reduce framing bias.

\vspace{2pt}

\noindent\textbf{GPT-3.5} is a large language model that generates abstractive summaries via prompting. We use gpt-3.5-turbo version and prompt is in Appendix \ref{gpt_prompt}.

\vspace{2pt}

\noindent\textbf{GPT-3.5 + one-shot} provides one example of three articles and their neutralized summary as a demonstration into the gpt-3.5-turbo model.

\vspace{2pt}

\noindent\textbf{GPT-3.5 + graph} guides the gpt-3.5-turbo model to firstly reason the event relation graph of the given articles, and then generate the summary through a chain-of-thought process \cite{wei2023chainofthought}.

\vspace{2pt}

\noindent\textbf{GPT-4} is another large language model that automatically generates abstractive summaries. We use the gpt-4 version to create the summaries.

\vspace{2pt}

\noindent\textbf{GPT-4 + graph} incorporates the graph into gpt-4 model through a chain-of-thought process.

\subsection{Automatic Evaluation}

The automatic evaluation aims to evaluate the summarization models from the perspective of content preservation and bias mitigation. To evaluate content preservation, we calculate Rouge scores \cite{lin-2004-rouge} and BLEU score \cite{papineni-etal-2002-bleu} between model generated summaries and human written reference. As bias can be induced by both emotional language and biased information included in the content \cite{fan-etal-2019-plain}, to evaluate bias mitigation, we evaluate both lexical-level bias and content-level informational bias, by calculating the arousal scores and polarization score respectively (calculation detailed in Appendix \ref{automatic_evaluation_appendix}). The arousal scores designed by \citet{lee-etal-2022-neus} include three metrics: positive-arousal, negative-arousal, and sum-arousal. A better summarizer is expected to attain higher Rouge and BLEU scores, as well as lower polarization score and arousal scores. The automatic evaluation results are shown in Table \ref{automatic_evaluation}.

The results demonstrate that incorporating the multi-document event relation graph effectively mitigates both lexical and informational bias, and meanwhile improving content preservation for both Llama-2 and LED models. Compared to their baseline models without the graph, our approach successfully alleviates media bias and reduces the polarization score and arousal scores. Besides, our approach notably improves content preservation and  yields higher Rouge and BLEU scores. In addition, incorporating the multi-document event relation graph effectively improves content preservation as well as alleviates media bias for commercial GPT models as well, reducing polarization and arousal scores and meanwhile yielding higher Rouge and BLEU scores. Moreover, our approach based on the multi-document event relation graph outperforms previous methods that lack bias information.
%This shows that the multi-document event relation graph captures event-level narrative structures and integrates bias information into LLMs, thereby improving content preservation as well as alleviating media bias.
%This proves that the multi-document event relation graph guides the LLMs to generate better balanced and unbiased summaries. Moreover, our approach based on the multi-document event relation graph outperforms previous methods that lack bias information. 
Overall, the finetuned Llama-2 + graph model (the very last row) achieves the highest Rouge scores, BLEU score, and the lowest polarization score, and the finetuned LED + graph model (the last row of LED models) exhibits the lowest arousal scores.
%This indicates that informing the LLMs of bias distribution is necessary for generating a neutralized summary. The multi-document event relation graph captures event-level narrative structures and integrates bias information into LLMs, thereby improving content preservation as well as alleviating media bias.

\begin{figure*}[t]
  \centering
  \includegraphics[width = 6.3in]{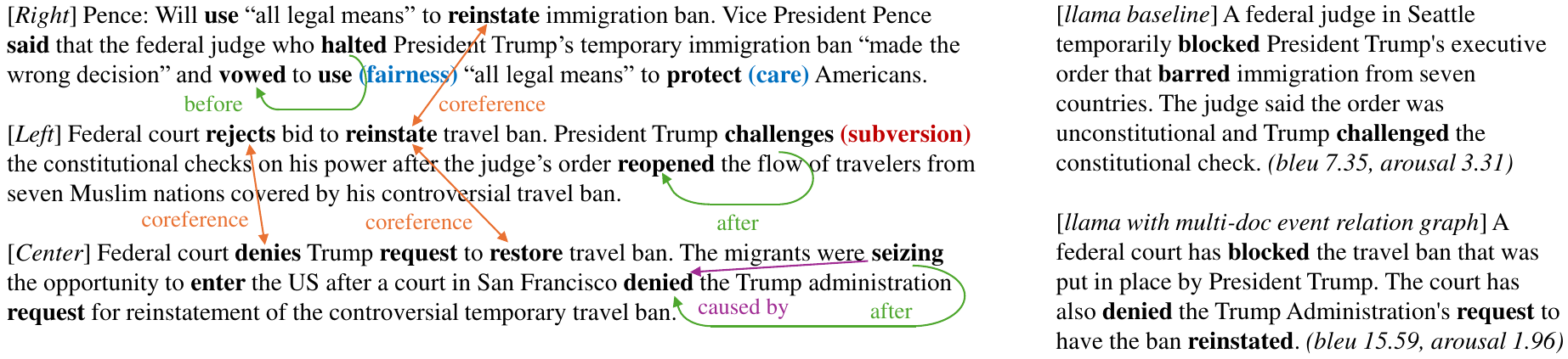}
  \caption{A qualitative analysis of the generated text before and after incorporating the multi-document event relation graph. The multi-document event relation graph effectively mitigates framing bias for neutralized summarization.}
  \label{analysis_example}
\end{figure*}

\subsection{Human Evaluation}

The human evaluation aims to evaluate the generated summaries from three perspectives: bias mitigation, content semantics, and language quality. Specifically, we design six questions regarding lexical bias (usage of emotional or biased words), informational bias (presentation of biased information that reflects ideological leanings), content non-hallucination, 
content recovery, language fluency, and language coherency. After collecting the evaluation scores from human annotators for the six questions, we normalize the metrics into the range of zero to one. For all the six metrics, a higher score is better. The details about human evaluation are presented in Appendix \ref{human_evalation_appendix}. The results of human evaluation are presented in Table \ref{human_evaluation}.

The human evaluation validates the consistent observations with automatic evaluation: the approach based on the multi-document event relation graph effectively mitigates both lexical bias and informational bias. This demonstrates that the multi-document event relation graph informs the LLMs of bias distribution, and guides the LLMs to utilize less biased words as well as present less biased information. Besides, the multi-document event relation graph also improves content quality, by reducing the hallucination and enhancing content recovery. The multi-document event relation graph explicitly show the reported events and represents the event-level content structures within and across documents, thereby assists in reducing hallucinations and recovering salient events.

Different from automatic evaluation, human evaluation indicates that GPT-4 model shows the best content quality among the evaluated models, with the least hallucination and best content recovery. One explanation is that Rouge and BLEU scores are calculated based on human written references, while human evaluation measures content non-hallucination and content recovery rate with respect to the input articles. Both automatic evaluation and human evaluation complement each other to achieve a comprehensive assessment of summary quality. Despite its strong performance in content quality, the GPT-4 model still suffers from a higher level of lexical bias and informational bias. This shows that the current powerful LLMs can still carry ideological bias in the generated content and mitigating ideological bias is necessary.

\subsection{Ablation Study}

The ablation study of the two designed strategies is shown in Table \ref{automatic_evaluation}. The results demonstrate that both textualized graph and graph prompt play a necessary role in mitigating media bias and improving content preservation. The textual graph and graph prompt complement each other: the textual graph augments the instruction prompt with graph information, while the graph prompt enables direct learning from the graph embedding. Incorporating the two strategies together achieves the best performance for both Llama-2 and LED models.

The ablation study of different components in the multi-document event relation graph is presented in Table \ref{ablation_study}. The results show that all the components in the graph are critical in mitigating media bias. The different elements in the graph carry complementary bias information:  event-level moral opinions inform opinionated moral bias,  within-doc event relations reflect different narrative framings of each article, and  cross-doc event coreference relation highlights event selection bias across various articles. All these components are %essential 
useful in constructing a unified and cohesive content structure. Integrating them together as a whole graph yields the best performance in terms of both content preservation and bias mitigation.

\subsection{Qualitative Analysis}

Figure \ref{analysis_example} shows a qualitative analysis of the generated summaries before and after incorporating the multi-document event relation graph into Llama-2 model (An example of the LED model is shown in Appendix \ref{qualitative_analysis_appendix}). The generated summary from the Llama-2 baseline (upper text) does not mention the core event \textit{denies the reinstatement}, but includes biased information \textit{challenged the constitutional check} from the liberal article that presents the liberal ideological bias, and also has hallucination \textit{a federal judge from Seattle} which contradicts with the input \textit{a court in San Francisco}. After incorporating the multi-document event relation graph, the generated summary (lower text) brings the mutually reported events \textit{denied the reinstatement} back, successfully excludes the biased information, and eliminates the hallucination. This shows the effectiveness of multi-document event relation graph in improving content preservation and mitigating media bias.

\section{Conclusion}

This paper aims to generate a neutralized summary given multiple articles with differing ideological bias as input, thereby mitigating their framing bias. Motivated by the critical roles of events and event relations in detecting media framing bias, we propose to build a multi-document event relation graph to inform LLMs of bias distribution. We further design two strategies to incorporate the multi-document event relation graph into LLMs for guiding the multi-document summarization process, which include graph textualization and graph prompt tuning. Both automatic evaluation and human evaluation demonstrate the effectiveness of our approach in mitigating media bias and meanwhile improving content preservation.

\section*{Limitations}

Our paper proposes to construct a multi-document event relation graph to guide the neutralized summarization process. The performance of this graph is not perfect and may still make errors in extracting event relations. We observe that the current event relation graph has the ability to extract event relations for most easy cases with explicit discourse connectives or language cues. However, it may make mistakes in recognizing hard cases that state event relations in an implicit way. To further improve the performance of media bias mitigation, it is necessary to enhance the extraction of implicit event relations. Therefore, improving the construction of multi-document event relation graph becomes necessary and serves as the future work.

\section*{Ethical Considerations}

This paper develops methodology to mitigate media bias. The media framing bias is a type of unwanted bias, which has harmful impact on both individuals and the society, such as misleading readers, intensifying societal polarization, and undermining democratic values. The goal of this paper is to mitigate the unwanted media framing bias and enhance unbiased information access. The examples in this paper are only used for research purpose, and do not represent any political leaning of the authors. The release of code and model should be leveraged to address and reduce media bias, serving a broader social good.

\section*{Acknowledgments}

We would like to thank the anonymous reviewers for their valuable feedback and input. We gratefully acknowledge support from National Science Foundation via the awards IIS-2127746 and IIS-1942918. Portions of this research were conducted with the advanced computing resources provided by Texas A\&M High-Performance Research Computing.

\bibliography{custom}

\appendix

\newpage

\section{Multi-document Event Relation Graph}
\label{eval_graph}

\subsection{Implementation Details}

An event identification model is trained on the MAVEN dataset which annotated event mentions for general-domain documents \cite{wang-etal-2020-maven}. Considering the news articles are usually long, we use the Longformer \cite{beltagy2020longformer} language model to encode the article and build a binary classification head on top of the word embeddings to predict whether the word triggers an event or not.

The event-level moral opinions classifier is trained based on the EMONA dataset which annotated moral opinions towards each event in news articles \cite{lei2024emona}. Following \citet{lei2024emona}, we use the Longformer \cite{beltagy2020longformer} to encode the entire article and add an extra layer of Bi-LSTM \cite{huang2015bidirectional} on top to capture the contextual information. Then we build a 11-class classification head on top of the event words embeddings to predict the moral label. The 11 labels include ten moral foundations (care, harm, fairness, cheating, loyalty, betrayal, authority, subversion, purity, degradation) and the non-moral label.

The four event relation extractors are trained on the general-domain MAVEN-ERE dataset \cite{wang-etal-2022-maven}. We follow previous work \cite{han-etal-2019-joint, yao2020weakly} to form the training event pairs in natural textual order, meaning the former event in the pair is the precedent event mentioned in text. For the temporal relations, the dataset also annotates the time expressions such as date or time. Considering our event relation graph focuses on events, we only retain annotations between events. We further process the \textit{before} annotation in this way: keep the \textit{before} label if the annotated event pairs aligns with the natural textual order, otherwise assign \textit{after} label to the reverse pair. The \textit{simultaneous}, \textit{overlap}, \textit{begins-on}, \textit{ends-on}, \textit{contains} annotations are grouped into \textit{overlap} category in our graph. For the causal relations, we keep the \textit{cause} label if the natural textual order is followed, or assign the \textit{caused by} label if not. For the \textit{subevent} relations, we maintain the \textit{contain} label if the natural textual order is followed, and assign \textit{contained by} label otherwise. The Longformer \cite{beltagy2020longformer} is used as the foundation language model, and the event pair embedding is the concatenation of two event words embeddings. Since the four event relations interact with each other to form a cohesive narrative structure, we adopt the joint learning framework \cite{wang-etal-2022-maven} to train these relations collaboratively.

\subsection{Evaluation Performance}

Table \ref{event_identification_result} presents the performance of event identification. Table \ref{event_moral_result} shows the performance of event-level moral opinions classification, where we use macro precision, recall, and F1 score as evaluation metrics. Table \ref{coreference_result} shows the performance of event coreference resolution. Following the previous work \cite{cai-strube-2010-evaluation}, MUC \cite{vilain-etal-1995-model}, $B^3$ \cite{10.3115/980845.980859}, $CEAF_e$ \cite{luo-2005-coreference}, and BLANC \cite{recasens_hovy_2011} are used as evaluation metrics. The performances of other components in the event relation graph, including temporal, causal, and subevent relation classification are summarized in Table \ref{temp_causal_subevent_result}. The standard macro-average precision, recall, and F1 score are reported.

\begin{table}[ht]
    \centering
    \scalebox{1.0}{
    \begin{tabular}{|c|ccc|}
        \hline
        & Precision & Recall & F1 \\
        \hline
        Event Identifier & 87.31 & 91.81 & 89.40 \\
        \hline
    \end{tabular}}
    \caption{Performance of event identification. Macro precision, recall, and F1 are reported.}
    \label{event_identification_result}
\end{table}

\begin{table}[ht]
    \centering
    \scalebox{1.0}{
    \begin{tabular}{|c|ccc|}
        \hline
        & Precision & Recall & F1 \\
        \hline
        Moral Classifier & 45.75 & 38.94 & 41.13 \\
        \hline
    \end{tabular}}
    \caption{Performance of event-level moral opinions classification. Macro precision, recall, and F1 are reported.}
    \label{event_moral_result}
\end{table}

\begin{table}[ht]
    \centering
    \scalebox{1.0}{
    \begin{tabular}{|c|ccc|}
        \hline
        & Precision & Recall & F1 \\
        \hline
        Temporal & 48.45 & 46.43 & 47.04 \\
        Causal & 58.48 & 54.02 & 56.01 \\
        Subevent & 53.37 & 42.90 & 46.21 \\
        \hline
    \end{tabular}}
    \caption{Performance of temporal, causal, and subevent relation tasks in the event relation graph. Macro precision, recall, and F1 are reported.}
    \label{temp_causal_subevent_result}
\end{table}

\begin{table*}[ht]
    \centering
    \scalebox{0.87}{
    \begin{tabular}{|ccc|ccc|ccc|ccc|}
        \hline
        \multicolumn{3}{|c|}{MUC} & \multicolumn{3}{|c|}{$B^3$} & \multicolumn{3}{|c|}{$CEAF_e$} & \multicolumn{3}{|c|}{BLANC} \\
        \hline
        Precision & Recall & F1 & Precision & Recall & F1 & Precision & Recall & F1 & Precision & Recall & F1 \\
        76.34 & 83.10 & 79.57 & 97.07 & 98.32 & 97.69 & 97.79 & 97.00 & 97.39 & 83.69 & 92.43 & 87.54 \\
        \hline
    \end{tabular}}
    \caption{Performance of event coreference resolution in the event relation graph}
    \label{coreference_result}
\end{table*}

\subsection{Statistical Analysis}

Table \ref{graph_statistics} presents the statistics of multi-doc event relation graph on NeuS dataset. On average, there are 25.59 events, 2.81 moral events, 1.82 coreference relation, 38.64 temporal relation, 6.27 causal relation, 1.24 subevent relation, and 3.42 cross-doc coreference relation within one graph.

\begin{table*}[ht]
    \centering
    \scalebox{0.9}{
    \begin{tabular}{|cccccccc|}
        \hline
        events & moral events & event pairs & coreference & temporal & causal & subevent & cross-doc coref \\
        \hline
        25.59	& 2.81 & 118.53	& 1.82	& 38.64	& 6.27	& 1.24	& 3.42 \\
        \hline
    \end{tabular}}
    \caption{The statistics of multi-document event relation graph on NeuS dataset. The average number of events, moral events, event pairs, four in-doc event relations, and cross-doc event coreference relation in a graph are shown.}
    \label{graph_statistics}
\end{table*}

\section{Instruction Prompt for Summarization}
\label{method_prompt}

The instruction prompt provided into Llama-2 and LED baseline models is: "Please summarize the given text. Text: <article 1> /s <article 2> /s <article 3>. Summary:"

The instruction prompt to incorporate the textualized graph into Llama-2 and LED models is: "The task is summarizing the given text. The events and event relations are important for summarization. An event is an occurrence or action reported in the text. The moral attribute of event represents the event is objective or contains subjective moral evaluation. The following table presents the events in the text, and the columns are event id, event word, and event moral attribute: <event table>. There are four types of event relations: coreference, temporal, causal, and subevent relations. Coreference relation represents two events designate the same occurrence. Temporal relation represents the chronological order between events, such as before, after, and overlap. Causal relation represents the causality between events. Subevent relation represents the containment relation from a parent event to a child event. The following table presents the event relations in the text, and the columns are source event id, source event word, relation between source event and target event, target event id, target event word: <event relation table>. Please summarize the given text. Text: <article 1> /s <article 2> /s <article 3>. Summary:"

\section{Prompt for GPT-based baselines}
\label{gpt_prompt}

The prompt provided into gpt-3.5-turbo abd gpt-4 baselines is: "Please summarize the given text. Text: <article 1> /s <article 2> /s <article 3>. Summary:"

The prompt provided into gpt-3.5-turbo + one-shot baseline is: "Please summarize the given text. Please mimic the output style in the following example. Example: <example article 1> /s <example article 2> /s <example article 3>. Summary: <example summary>. Text: <article 1> /s <article 2> /s <article 3>. Summary:"

The prompt provided into gpt-3.5-turbo + graph baseline is: "Please summarize the given text. Let's think step by step. Firstly, explain the events reported in each article and the relations between events. Events refer to an occurrence or action reported in the sentence. There are four types of event relations: coreference, temporal, causal, and subevent relations. Coreference relation represents two events designate the same occurrence. Temporal relation represents the chronological order between events, such as before, after, and overlap. Causal relation represents the causality between events. Subevent relation represents the containment relation from a parent event to a child event. Secondly, generate the summary. Please mimic the output style in the following example. Example: <example article 1> /s <example article 2> /s <example article 3>. Output: Firstly, explain the events reported in each article and the relations between events. <explanation of events and event relations in the example articles> Secondly, generate the summary. Summary: <example summary>. Text: <article 1> /s <article 2> /s <article 3>. Output:"

\section{Automatic Evaluation}
\label{automatic_evaluation_appendix}

The automatic evaluation aims to evaluate the summarization models from the perspective of content preservation and bias mitigation.

For the content evaluation, we use Rouge scores (Rouge-1, Rouge-2, Rouge-L, Rouge-Lsum scores) and BLEU score (cumulative BLEU-2 score) between model generated summaries and human written reference as metrics \cite{lin-2004-rouge, papineni-etal-2002-bleu}. The Rouge scores are implemented by using the rouge\_scorer.RougeScorer function from the rouge\_score package. The cumulative BLEU-2 score is implemented by using the sentence\_bleu function from the nltk.translate.bleu\_score package, with a weight of (0.5, 0.5, 0, 0).

For the bias evaluation, we evaluate both lexical-level bias and content-level informational bias, by using the arousal scores and polarization score respectively. This is because bias can be induced by both emotional language and biased information within content. The arousal scores are designed by \citet{lee-etal-2022-neus} and evaluate lexical-level emotion. Specifically, the arousal scores are computed based on the Valence-Arousal-Dominance emotion lexicon \cite{mohammad2018obtaining} and include three metrics: p-arousal (arousal scores of positive words in the summary), n-arousal (arousal scores of negative words), and sum-arousal (summation of positive and negative arousal). On the other hand, the polarization score evaluates content-level informational bias. To calculate polarization score, an ideology classifier is trained on the AllSides dataset \cite{baly2020we}, where a classification head is built on top of the BERT language model to classify article-level ideology into liberal, center, and conservative. The macro F1 score of this ideology classifier is 80.19, which proves its validity. The polarization score is computed as the predicted probability of the polarized class, i.e. the sum of predicted probability of liberal and conservative classes. The polarization score ranges from 0 to 1, with 1 indicating full ideological polarization (liberal or conservative) and 0 indicating no polarization (neither liberal nor conservative).

\begin{table*}[ht]
    \centering
    \scalebox{0.9}{
    \begin{tabular}{|l|c|}
        \hline
        & generated summary \\
        \hline
        golden summary & \makecell[l]{Travel Ban Reinstatement Rejected. A federal court rejected the reinstatement of the \\travel ban. The migrants from the seven listed countries are allowed to enter the US.} \\
        \hline
        LED & \makecell[l]{Vice President Mike Pence said Sunday that the federal judge who halted President \\Trump’s temporary immigration ban “made the wrong decision” and vowed to use\\ “all legal means at our disposal” to protect Americans.} \\
        \hline
        LED + graph & \makecell[l]{Appeals Court Denies Request to Immediately Reinstate Travel Ban. A federal appeals\\ court denied the Trump administration's request to immediately reinstate the travel ban.} \\
        \hline
        Llama-2 & \makecell[l]{A federal judge in Seattle temporarily blocked President Trump's executive order that\\ barred immigration from seven countries. The judge said the order was unconstitutional\\ and Trump challenged the constitutional check.} \\
        \hline
        Llama-2 + graph & \makecell[l]{A federal court has blocked the travel ban that was put in place by President Trump. \\The court has also denied the Trump Administration's request to have the ban reinstated.} \\
        \hline
    \end{tabular}}
    \caption{A qualitative example of the generated summaries before and after incorporating the multi-document event relation graph into LED and Llama-2 models.}
    \label{generated_summary_example}
\end{table*}

\section{Human Evaluation}
\label{human_evalation_appendix}

The human evaluation aims to evaluate the generated summaries from three perspectives: bias mitigation, content semantics, and language quality. Considering bias can be induced by both lexical usage and information selection \cite{fan-etal-2019-plain}, we evaluate both types of lexical bias (usage of emotional or biased words) and informational bias (presentation of biased information that reflects ideological leanings). Specifically, we design six questions regarding lexical bias, informational bias, content non-hallucination, content recovery, language fluency, and language coherency. The detailed questions are:

\begin{enumerate}
    \item Question (lexical \textbf{bias}): Does the text use biased or emotional words? Choose 2, 1, 0. Score Explanation: 2 - uses objective and neutral words, 1 - uses some emotional words but acceptable, 0 - uses biased and emotional words.
    \item Question (informational \textbf{bias}): Does the text contain biased or polarized information to showcase ideology leaning? Choose 2, 1, 0. Score Explanation: 2 - does not showcase ideology leaning, 1 - conveys some ideology bias but acceptable, 0 - contains biased information to showcase ideology leaning.
    \item Question (\textbf{content} non-hallucination): Does the text hallucinate compared to input articles? Choose 1 or 0. Score Explanation: 1 - does not hallucinate and aligns with the facts in input articles, 0 - hallucinates some content that is not included in the input articles.
    \item Question (\textbf{content} recovery): Does the text recover the important content from input articles? Choose 1 or 0. Score Explanation: 1 - successfully recover important content from input articles, 0 - misses some important content in input articles.
    \item Question (\textbf{language} fluency): Is the text fluent and grammarly correct? Choose 1 or 0. Score Explanation: 1 - fluent and grammarly correct, 0 - not fluent and has grammar errors.
    \item Question (\textbf{language} coherency): Is the text coherent with natural logic flow? Choose 1 or 0. Score Explanation: 1 - coherent and logic flow is natural, 0 - not coherent and logic flow is not natural.
\end{enumerate}

There are two human annotators participated in the evaluation, both are graduate students who are familiar with natural language processing and media bias research. Their reported political leanings are center. One annotator is native English speaker and the other is proficient in English. Both the annotators were paid. The Cohen's kappa inter-annotator agreement rate is 0.67. The randomly sampled 50 clusters of articles from the test set were annotated. We select the NeuS and GPT-4 baselines, LED, LED + graph (the last row of LED models in Table \ref{automatic_evaluation}), Llama-2, Llama-2 + graph (the very last row in Table \ref{automatic_evaluation}) for evaluation. To avoid the leakage of model information, different models are randomly shuffled and the name of models are omitted. After collecting the evaluation scores from human annotators for the six questions, we normalize the metrics into the range of zero to one. For all the six metrics, a higher score is better.

\section{Qualitative Analysis}
\label{qualitative_analysis_appendix}

Table \ref{generated_summary_example} presents an example of the generated summaries from LED and LED + graph models, as well as the summaries from Llama-2 and Llama-2 + graph models. The multi-document event relation graph alleviates the ideological leaning in the summaries for both LED and Llama-2 models.

\end{document}